\documentclass[10pt,onecolumn,letterpaper]{article}

\pdfoutput=1

\usepackage{cvpr}
\usepackage{times}
\usepackage{epsfig}
\usepackage{graphicx}
\usepackage{amsmath}
\usepackage{amssymb}

\usepackage[pagebackref=true,breaklinks=true,colorlinks,bookmarks=false]{hyperref}

\cvprfinalcopy % *** Uncomment this line for the final submission

\def\cvprPaperID{9753} % *** Enter the CVPR Paper ID here

\ifcvprfinal\fi
\usepackage{amsmath}
\usepackage{dsfont}

\usepackage{graphicx}
\usepackage{subfig}

\newcommand{\flops}{\mbox{flops}}

\usepackage{xcolor}
\newcommand{\red}[1]{\textcolor{black}{#1}} %comment this to remove red
\newcommand{\blue}[1]{\textcolor{black}{#1}} %comment this to remove red
\usepackage{setspace}
\begin{document}

%%%%%%%%% TITLE
\title{Interleaved Composite Quantization for High-Dimensional Similarity Search}

\author{Soroosh Khoram\\
Department of Electrical and Computer Engineering\\
%University of Wisconsin-Madison\\
%Madison, WI, USA\\
%{\tt\small khoram@wisc.edu}
\and
Steven J. Wright\\
Department of Computer Sciences\\
\and
Jing Li\\
Department of Electrical and Computer Engineering\\
University of Wisconsin-Madison\\
Madison, WI, USA\\
{\tt\small jli@ece.wisc.edu}
}

\maketitle

\begin{abstract}
Similarity search retrieves the nearest neighbors of a query vector from a dataset of high-dimensional vectors. As the size of the dataset grows, the cost of performing the distance computations needed to implement a query can become prohibitive. A method often used to reduce this computational cost is quantization of the vector space and location-based encoding of the dataset vectors. \red{These encodings} can be used during query processing to find {\em approximate} nearest neighbors of the query point quickly. \blue{Search speed can be improved by using shorter codes, but shorter codes have higher quantization error, leading to degraded precision. In this work, we propose the {\em Interleaved Composite Quantization} (ICQ) which achieves fast similarity search without using shorter codes.} In ICQ, a small subset of the code is used to approximate the distances, with complete codes being used only when necessary. Our method effectively reduces both code length and quantization error. Furthermore, ICQ is compatible with several recently proposed techniques for reducing quantization error and can be used in conjunction with these other techniques to improve results. We confirm these claims  and show strong empirical performance of ICQ using several synthetic and real-word datasets.
%% , showing that the proposed method can outperform recent, state-of-the-art quantization techniques.

% Many machine learning algorithms generate datasets of high-dimensional vectors. A common operation performed on such datasets is similarity search, where we search for the nearest neighbors in the dataset of a given query point. This operation is expensive when performed naively and repeatedly. Numerous methods have been proposed to reduce this computational cost, often by quantizing the vector space and encoding the vectors in the dataset based on their locations. This coding is used during query processing to find the approximate nearest neighbors of the query point quickly. However, faster searches in these methods rely on more aggressive quantizations and shorter codes, which can degrade precision due to the higher quantization error. In this work, we aim to use less aggressive quantization and accept longer codes without sacrificing the search speed. In the proposed encoding, a small subset of the code can be used to approximate the distances, with complete codes being used only when necessary. Our method effectively reduces both code length and quantization error. Furthermore, the proposed technique is compatible with several recently proposed techniques for reducing quantization error and can be used in conjunction with them to improve results. We confirm these claims empirically using several synthetic and real-word datasets, showing that the proposed method can outperform recent, state-of-the-art quantization techniques.
\end{abstract}

\section{Introduction}
Similarity search, which finds objects in a dataset that are similar to a query object, is a common operation in machine learning. Similarity is often quantified via a straightforward distance computation between embeddings of the objects in a vector representation space. Implementation of similarity search becomes computationally expensive when information explosion and the curse of dimensionality conspire to make exhaustive search impractical. In this case, hashing and Vector Quantization (VQ) are often used to simplify and streamline the computations. As VQ methods generally outperform hashing techniques \cite{wu2017multiscale}, our focus is on this class of similarity search techniques.

% For a dataset $\mathbf{X}=\{\mathbf{x}_1, \cdots, \mathbf{x}_n\}\subset\mathbb{R}^d$, VQ learns a quantizer $\mathcal{C}=\{\mathbf{c}_1,\cdots,\mathbf{c}_m\}\subset\mathbb{R}^d$ and simplifies the search by assigning each $\mathbf{x}_i$ to an element of the quantizer, denoted by $\overline{\mathbf{x}}_i\in\mathcal{C}$. A query $\mathbf{q}\in\mathbb{R}^d$ needs only to be compared against $\mathcal{C}$ rather than to all of $\mathbf{X}$, to approximately find its nearest neighbors. Normally, however maintaining a tolerable quantization error would require an impractically large value of $m$. In practice, VQ methods learn $K$ quantizers, denoted by $\mathcal{C}_k=\{\mathbf{c}_{k,1}, \cdots, \mathbf{c}_{k,m}\}$ for $k=1,2,\dotsc,K$, and combine them to reconstruct an approximation of the dataset. A dataset element $\mathbf{x}_i$ is usually quantized to a {\em sum} of elements of the quantizers, that is, $\overline{\mathbf{x}}_i=\sum_{k=1}^K\overline{\mathbf{x}}_{k,i}$, where $\overline{\mathbf{x}}_{k,i}\in\mathcal{C}_k$. This composition allows us to encode $\mathbf{x}_i$. \red{This approach} allows for reduction of the quantization error with smaller quantizer sets and compression of the dataset through a lossy encoding of embeddings, and given certain properties for $\mathcal{C}$, we have
For a dataset $\mathbf{X}=\{\mathbf{x}_1, \cdots, \mathbf{x}_n\}\subset\mathbb{R}^d$, VQ learns a quantizer $\mathcal{C}=\{\mathbf{c}_1,\cdots,\mathbf{c}_m\}\subset\mathbb{R}^d$ and simplifies the search by assigning each $\mathbf{x}_i$ to an element of the quantizer, denoted by $\overline{\mathbf{x}}_i\in\mathcal{C}$. A query $\mathbf{q}\in\mathbb{R}^d$ needs only to be compared against $\mathcal{C}$ rather than to all of $\mathbf{X}$, to approximately find its nearest neighbors. Normally, however maintaining a tolerable quantization error would require an impractically large value of $m$. In practice, VQ methods learn $K$ quantizers, denoted by $\mathcal{C}_k=\{\mathbf{c}_{k,1}, \cdots, \mathbf{c}_{k,m}\}$ for $k=1,2,\dotsc,K$, and combine them to reconstruct an approximation of the dataset. A dataset element $\mathbf{x}_i$ is usually quantized to a {\em sum} of elements of the quantizers, that is, $\overline{\mathbf{x}}_i=\sum_{k=1}^K\overline{\mathbf{x}}_{k,i}$, where $\overline{\mathbf{x}}_{k,i}\in\mathcal{C}_k$. \red{This approach} allows for reduction of the quantization error with smaller quantizer sets and compression of the dataset through a lossy encoding of embeddings, and given certain properties for $\mathcal{C}$, we have
\begin{equation}
    \label{eq:vq_comparison}
    \sum_{k=1}^K \lVert \mathbf{q} - \overline{\mathbf{x}}_{k,i}\rVert^2 < \sum_{k=1}^K \lVert \mathbf{q} - \overline{\mathbf{x}}_{k,i^\prime} \rVert^2 \Rightarrow \lVert \mathbf{q}-\overline{\mathbf{x}}_i \rVert^2 < \lVert \mathbf{q}-\overline{\mathbf{x}}_{i^\prime} \rVert^2.
\end{equation}
% \blue{This principle is often used for fast similarity search. Since $\overline{\mathbf{x}}_{k,i}\in\mathcal{C}_k$, we can precompute all values of $\lVert \mathbf{q}-\mathbf{c}_{k,i}\rVert$ and, during search time, perform only $K$ summations for each dataset element $\overline{\mathbf{x}}_i$. Faster searches in existing works are often achieved by reducing $K$. \red{This approach} can increase quantization error, so techniques are proposed to minimize this effect, for example, jointly learning the embedding method and the quantizers (Section~\ref{sec:related}). Nevertheless, as aggressive quantization necessarily sacrifices information, requirements of search accuracy place an implicit lower bound on $K$, thus limiting maximum attainable search speed.}

\blue{This principle is often used for fast similarity search. Since $\overline{\mathbf{x}}_{k,i}\in\mathcal{C}_k$, we can precompute all values of $\lVert \mathbf{q}-\mathbf{c}_{k,i}\rVert$ and, during search time, perform only $K$ summations for each dataset element $\overline{\mathbf{x}}_i$. Faster searches in existing works are often achieved by reducing $K$. However, reducing $K$ can increase quantization error, so techniques are proposed to minimize this effect, for example, jointly learning the embedding method and the quantizers (Section~\ref{sec:related}). Nevertheless, as aggressive quantization necessarily sacrifices information, requirements of search accuracy place an implicit lower bound on $K$, thus limiting maximum attainable search speed.}

\blue{In this work, we propose {\em Interleaved Composite Quantization} (ICQ), a scheme that enables fast searches without the need of reducing $K$. ICQ dedicates a subset of the quantizers, say $\mathcal{K}\subset[1,K]$, to fast, approximate distance comparisons. This is done by learning a probabilistic model of the distribution of the dataset which guides clustering of the quantizers such that the following equation implies that $\overline{\mathbf{x}}_i$ is potentially closer to $\mathbf{q}$ than $\overline{\mathbf{x}}_{i^\prime}$.}
\begin{equation}
    \sum_{k\in\mathcal{K}} \lVert \mathbf{q} - \overline{\mathbf{x}}_{k,i}\rVert^2 < \sum_{k\in\mathcal{K}} \lVert \mathbf{q} - \overline{\mathbf{x}}_{k,i^\prime} \rVert^2 + \sigma
    \label{eq:ssc_comparison}
\end{equation}
\blue{In this comparison, the margin $\sigma$ accounts for the variability of the remaining quantizers. We refine distance comparisons using equation \ref{eq:vq_comparison} only when necessary. Since in many real-world cases, most dataset elements are far more distant from a random query than its nearest neighbors, the comparison \eqref{eq:ssc_comparison} suffices to prune neighbor candidates, reducing the number of operations to $|\mathcal{K}|<K$ for most dataset elements compared to $K$ for the previous works.}

\blue{ICQ is the first similarity search method based on crude distance comparisons. Search speed improvement of ICQ over previous works depends critically on two key contributions. First, we {\em learn} the probabilistic model of the dataset distribution that is used to cluster the quantizers for fast distance comparisons. Both the learning of this model and the clustering can be performed jointly with the learning of the embedding method and the quantizers. As such, ICQ can benefit from joint methods proposed in the literature for improving quantization error (Section~\ref{sec:related}). Second, we propose a new two-step similarity search method, in which we first perform fast, crude distance comparisons and then refine these comparisons only when necessary. In this way, we perform fast searches without increasing quantization error.}

\section{Related Works}
\label{sec:related}
Before gaining prevalence in similarity search, VQ methods were studied as a form of lossy encoding \cite{gray1990vector,gersho2012vector} for compression. \cite{jegou2010product} first introduced Product Quantization (PQ) for nearest neighbor search with asymmetric distance comparison (equation \ref{eq:vq_comparison}). In this method, each of $\mathcal{C}_k$ quantizes one of $K$ predetermined, orthogonal subspaces of $\mathbb{R}^d$, comprising $\frac{d}{K}$ consecutive dimensions, i.e. its elements had exactly $\frac{d}{K}$ consecutive non-zero elements. The quantizer for each subspace was learned using k-means on the projection of the dataset onto it. 

Since search speed in PQ methods has been often determined by the size of the quantizers, extensions of this method have tried to reduce the quantization error using fewer quantizers. \cite{ge2013optimized} proposed Optimized PQ (OPQ) which learns a rotation matrix that better reorders the dimensions and aligns the dataset with the PQ subspaces. Concurrently, \cite{norouzi2013cartesian} proposed orthogonal k-means which combines a similar rotation matrix and a translation vector. Locally Optimized PQ further reduced quantization error by learning several rotation matrices that are locally optimized and is learned jointly with the quantizers. Norouzi et al. achieved a similar effect as the rotation methods by learning quantizers orthogonal to each other but not necessarily aligned with the dimensions of the embedding space.

Additive methods further relaxed the constraints on learning quantizers by allowing all quantizers to span across all of $\mathbb{R}^d$. \cite{zhang2014composite} proposed Composite Quantization (CQ) which required quantizers only to have a constant inner product instead of being orthogonal. Supervised Quantization \cite{wang2016supervised} jointly learned CQ quantizers and a linear mapping for embedding the dataset, combining the embedding and quantization.

Recently, there has been several methods that jointly learn embedding models and quantizers. \cite{klein2017defense} proposed jointly learning a DNN model and PQ quantizers in an end-to-end manner by adding a fully connected layer that learns quantizer assignments. PQN \cite{yu2018product} similarly learns an end-to-end network but removes the need for a fully connected layer.

\blue{Unlike previous works which trade quantization error for speed. ICQ instead, by clustering the quantizers and fast, crude distance comparisons based on equation \ref{eq:ssc_comparison} avoids reducing the number of quantizers and increasing the quantization error. Furthermore, as the key components of ICQ, that is the probabilistic model of dataset distribution and the clustering of the quantizers (section 1) can be seamlessly integrated into the quantization process, it can benefit from many existing methods for further reducing quantization error, e.g. jointly learning the embeddings and the quantizers.}

% The main focus in the above mentioned approaches has been maintaining low quantization error when fewer quantizers are learned. Faster search time is achieved as a result of reducing the size of the quantizers. In this work, we take an alternative approach focusing on fast but crude distance comparisons. As such, we do not need a hard limit on the total number of the quantizers which frees us to use more quantizers to reduce quantization error.

\section{Methodology}
Fast similarity search through quantization requires two steps. First, the raw dataset $\mathcal{D}$ is mapped into the semantically meaningful set of embeddings $\mathbf{X}$ using the model $\mathbf{W}$. This is done by minimizing the error term $\mathcal{L}^E$, which can be defined as the classification loss, the triplet loss, or a similar measure of accuracy. Second, the quantizers $\mathcal{C}=\{\mathcal{C}_1,\cdots,\mathcal{C}_K\}$ are learned such that quantization error $\mathcal{L}^C$ is minimized. Still, as quantization can work against search accuracy, it is often advantageous to combine the two steps \cite{klein2017defense,wu2017multiscale,yu2018product,wang2016supervised} and perform quantization-aware embedding. This is equivalent to solving problem of equation \ref{eq:joint_loss} below. We note that, while the overall loss in these methods is a weighted sum of accuracy loss and quantization loss, since these two losses may comprise multiple, differently weighted terms themselves, here we assume these weights are included in the definitions.
\begin{equation}
    \min_{\mathbf{W}, \mathcal{C}} \mathcal{L}^E(\mathcal{D}, \mathbf{W})+\mathcal{L}^C(\mathbf{X}, \mathcal{C}).
    \label{eq:joint_loss}
\end{equation}
% \sjwcomment{Do you solve a weighted version of this where e.g. the second term is assigned some weight?  Or are the functions scaled to achieve the desired effect?}
In order for a subset of the quantizers ($\mathcal{K}$) to reliably compare proximities, they need to capture a substantial portion of the variability of the data. We realize this in two steps. First, we identify a subspace of $\mathbb{R}^d$ with high dataset variance. Then, we cluster quantizers such that a small subset of them are dedicated to quantizing that subspace. To identify the high-variance subspace we incorporate the variances of the $\mathbf{X}$, $\mathbf{\Lambda}=[\lambda_1, ..., \lambda_d]^T\in\mathbb{R}_{\geq 0}^d$ into our model training and quantization. Normally, in the real-world data, there is a high variance in the distribution of $\mathbf{\Lambda}$ itself \cite{kalantidis2014locally}. We use the largest values of $\mathbf{\Lambda}$ to identify the high-variance subspace and allocate a few of the quantizers to it. In the rest of this section, we define an augmented loss function for this purpose, describe the learning procedure using this loss function, and discuss its features.

\subsection{Augmented Loss function}
We define a loss function based on the distribution of the variances $\mathbf{\Lambda}$. Variances in real-world data often follow a multi-modal distribution which we approximate using the prior $\mathbb{P}(\mathbf{\Lambda})$ defined as below:
\begin{align*}
    \mathbb{P}(\mathbf{\Lambda}; \mathbf{\Theta})=\prod_{i=1}^d\mathbb{P}(\lambda_i) = \prod_{i=1}^d\;\pi_1\mathcal{N}(\lambda_i;0,\sigma_1)+\\\pi_2\mathcal{SN}(\lambda_i;\mu_2, \sigma_2,\alpha_2)
    \label{eq:bi_modal} 
\end{align*}
This prior comprises a bi-modal mixture model. The first, major mode represented using the normal distribution $\mathcal{N}(.; 0, \sigma_1)$ centered around zero to allow for pruning of redundant features. We rely on the classification loss to prevent aggressive pruning of informative features. The second, minor mode is represented using the skew normal distribution $\mathcal{SN}(.; \mu_2, \sigma_2,\alpha_2)$ with a fixed, negative $\alpha_2$, since its asymmetry would attract $\lambda_i$ towards higher values. By choosing a higher mixing proportion for the major mode ($\pi_1 > \pi_2$), we encourage only a few high value variances. The parameter $\mathbf{\Theta}$ denotes the set of trinable parameters of this distribution: $\{\sigma_1, \sigma_2, \mu_2\}$. Then, we only need to minimize the negative log likelihood of this distribution:
\begin{equation}
    \mathcal{L}^P(\mathbf{\Lambda}, \mathbf{\Theta}) = -\log(\mathbb{P}(\mathbf{\Lambda}, \mathbf{\Theta}))
\end{equation}
Minimizing this loss, identifies a small subspace $\psi\subset\mathbb{R}^d$ corresponding to the most high-variance dimensions. Specifically, for $e_j$ being the one-hot base vector of $\mathbb{R}^d$ in the $j$-th direction, we define
\begin{equation}
    \psi = \text{span}(\{\mathbf{e}_i|\pi_2\mathcal{SN}(\lambda_i)>\pi_1\mathcal{N}(\lambda_i)\})
\end{equation}
We further wish to similarly cluster the quantizers into two groups. A few quantizers, the set $\mathcal{K}$, should be allocated to quantize $\mathbf{X}$ mapped to $\psi$. These will be used for fast distance comparisons. The remaining quantizers are used for the complementary subspace to reduce quantization error. Consequently, quantizers in the two groups are orthogonal. We formalize this condition below:
\begin{align*}
    \forall i\in[1,d], j\in\mathcal{K},k\in\overline{\mathcal{K}}: \forall \mathbf{c}_1\in\mathcal{C}_j, \mathbf{c}_2\in\mathcal{C}_k:\\ (\mathbf{c}_1^T\mathbf{e}_i)(\mathbf{c}_2^T\mathbf{e}_i)=0
\end{align*}
This condition is similar to the PQ method where perpendicularity of quantizers is achieved by zeroing out elements. However, unlike in PQ where zero elements are consecutive, here their location are flexibly chosen by the algorithm, interleaved among the non-zero elements. Because of this, we refer to this method of quantization the Interleaved Composite Quantization (ICQ). For simplicity of the optimization, we sum the above set of equations into the condition below:
\begin{equation}
    \mathcal{L}^{ICQ}(\mathcal{C}, \mathbf{\xi})=\sum_{i=1}^m\sum_{\mathbf{c}\in\mathcal{C}_i}\lVert \mathbf{c}\circ\mathbf{\xi}\rVert\lVert \mathbf{c}\circ(\mathds{1}_d-\mathbf{\xi})\rVert=0
\end{equation}
Where, $\circ$ indicates element-wise multiplication, $\mathds{1}_d$ is a vector of size $d$ with all $1$ elements and $\mathbf{\xi}\in\mathbb{R}^d$ is defined as:
\begin{equation}
    \mathbf{\xi}_i=\begin{cases}
    1& \mathbf{e}_i\in\psi \\
    0& \mathbf{e}_i\in\overline{\psi}
    \end{cases}
\end{equation}
We use $\mathcal{L}^{P}$ and $\mathcal{L}^{ICQ}$ as a soft constraints in learning the quantizers. While this might not fully satisfy the original constraint, it is sufficient. That is because we ultimately need crude estimations of distance. Based on this, we learn the ICQ quantizers following the below optimization:
\begin{align*}
    \min_{\mathbf{W}, \mathcal{C}, \mathbf{\Theta}} \mathcal{L}^E(\mathcal{D}, \mathbf{W})+\mathcal{L}^C(\mathbf{X}, \mathcal{C})+\\\gamma_1\mathcal{L}^P(\mathbf{\Lambda}, \mathbf{\Theta})+\gamma_2\mathcal{L}^{ICQ}(\mathcal{C}, \mathbf{\xi})
\end{align*}
The two terms $\mathcal{L}^E$ and $\mathcal{L}^C$ are still necessary to ensure accuracy and quantization error. Furthermore, here $\mathbf{\Theta}$ is the concatenation of $\mathbf{\theta}_i$, the trainable parameters of $\mathbb{P}(\mathbf{\Lambda})$. After the quantization, the set of quantizers $\mathcal{K}$ is obtained according to the definition below.
\begin{equation}
    i\in \mathcal{K} \Longleftrightarrow \forall \mathbf{c}\in\mathcal{C}_i: \lVert \mathbf{c}\circ (\mathds{1}_d-\mathbf{\xi})\rVert < \lVert \mathbf{c}\circ\mathbf{\xi}\rVert
\end{equation}
\subsection{Optimization}
Optimizing the embedding model together with the quantizers can be a difficult task due to the complex relationship of the parameters for which we optimize. As such, early approaches opted to these tasks separately \cite{wang2016supervised}. Recent methods have facilitated this process by allowing optimization of the loss through gradient descent methods based on batch learning \cite{klein2017defense,yu2018product}. ICQ can use both approaches with respect to the model and quantizer parameters. The main difference in optimization for ICQ is the addition of optimization for prior parameters and computation of variances. 

Optimization for $\mathbf{\Theta}$ can be mainly performed using Expectation Maximization (EM) or gradient descent. EM methods exist that can be applied to Skew Normal distributions \cite{Tsung2007finite}. For simplicity of the optimization, we choose to use gradient methods. 

Computation of the $\mathbf{\Lambda}$ normally requires computation of all $\mathbf{X}$ using the model parameters $\mathbf{W}$. This can be impractically expensive since $\mathbf{W}$ and as a result $\mathbf{X}$ change constantly during training and is incompatible with batch learning. To get around this issue, for each batch in an epoch, we estimate the dataset variance using the variances of all the batches observed thus far. We compute this estimate using online variance computation as shown below:
\begin{align*}
    \mathbf{\Lambda}_b^{dataset} = \mathbf{\Lambda}_{b-1}^{dataset}+\frac{1}{b}(\mathbf{\Lambda}_b^{batch}-\mathbf{\Lambda}_{b-1}^{dataset})+\\\frac{1}{b}(1-\frac{1}{b})(\mathbf{\text{M}}_b^{batch}-\mathbf{\text{M}}_{b-1}^{dataset})^2    
\end{align*}
\begin{equation}
    \mathbf{\text{M}}_b^{dataset}=\mathbf{\text{M}}_{b-1}^{dataset}+\frac{1}{b}(\mathbf{\text{M}}_b^{batch}-\mathbf{\text{M}}_{b-1}^{dataset})
\end{equation}
Here, $\mathbf{\Lambda}_b^{dataset}$ and $\mathbf{\text{M}}_b^{dataset}$ are estimates of the variance and mean of the dataset up to batch number $b$ and $\mathbf{\Lambda}_b^{batch}$ and $\mathbf{\text{M}}_b^{batch}$ are the sample variance and mean of this batch. This online computation of variance requires a small sample variance computation, each time and is thus very fast and requires almost no additional memory. Furthermore, as we gather more information about the dataset in each epoch, we improve our estimate of the dataset variance.

\subsection{Learning Robustness}
While most parameters in the proposed loss are learned, there are three key parameters that we have chosen to fix, and our choice of these values can affect the outcome of the optimization. These parameters are: $\alpha_2$, $\pi_1$, and $\pi_2$. 

The parameter $\alpha_2$ controls the skewness of the skew normal function. Large absolute values thus can essentially create a half-normal distribution. However, our only requirement is that the skew normal mode is roughly $\max(\Lambda)$. This can be satisfied so long as this mode is sufficiently asymmetrical by setting the value of $\alpha_2=-10$, for example. 

Conversely, the loss can be sensitive to the weights $\pi_1$ and $\pi_2$. That is because the case where all variances $\lambda_i$ belong to only one of the modes is in the feasible set. If all fall into the one mode, the method tries to minimize or maximize all variances and introduce noise to the data. We get around this issue by adding the probability of the minor mode to the loss and then choosing a small $\pi_2$.
% If all variances fall into the minor mode, it tries to maximize variance in all directions and introduces noise. We get around this issue by adding the probability of each of the minor modes to the loss as well and then choosing a very small $\pi_2$ and minimize:
\begin{equation}
\begin{split}
    \mathcal{L}^P(\mathbf{\Lambda})&=-\log(\mathbb{P}(\mathbf{\Lambda}))-\log(\mathbb{P}(\mathcal{SN})) \\
    \mathbb{P}(\mathcal{SN}) &= \sum_{i=1}^{d} \pi_2\mathcal{SN}(\lambda_i) \\
\end{split}
\end{equation}
The new component in $\mathcal{L}^P$ is the probability of the second mode and is added for robustness. Adding this component guarantees that the second mode is not emptied out to delete useful information.

\subsection{Search Operation}
The search process is a key differentiator of the proposed method and existing techniques. The search operation divides the normal search process into the two steps of approximate distance comparisons and accurate comparisons. In the conventional search, we normally maintain a list of the $K$ nearest neighbors to a query and update this list while sifting through the dataset. If a new dataset element is closer to the query than the furthest element in the list, i.e. satisfies equation \ref{eq:vq_comparison}, it replaces that element in the list. While in concept this process is serial, it can be parallelized  efficiently \cite{zhang2018efficient,faiss}.

We also maintain a list of the nearest neighbors, but when testing a new dataset element, we first perform fast distance comparisons with the furthest element according to equation \ref{eq:ssc_comparison}. If this equation is satisfied, we perform exact comparisons according to equation \ref{eq:vq_comparison}. During the crude distance comparisons, the parameter $\sigma$ in equation \ref{eq:ssc_comparison} accounts for the distance uncertainty of the two dataset elements in the subspace $\overline{\psi}$. Thus, we use the variance of the dataset in this subspace.
\begin{equation}
    \sigma \sim \sum_{i\in\overline{\mathcal{K}}}\lambda_i.
\end{equation}
\section{Evaluations}
We evaluate the performance of our ICQ  method on both synthetic and real-world datasets, and compare with prominent methods proposed in the literature, specifically, Supervised Quantization (SQ) \cite{wang2016supervised} and Product Quantization Network (PQN) \cite{yu2018product}. SQ uses a linear mapping for embedding and Composite Quantization (CQ) for quantization, while PQN --- the current state of the art --- uses CNN for embedding and PQ for quantization. For comparison with each method, we learn a similar embedding method and replace the quantization step with ICQ. For these comparisons, we use the widely used Mean Average Precision (MAP) and show that the proposed method generally is able to outperform existing methods.

\subsection{Setup}
We use both synthetic and real-world datasets to verify the performance of the proposed method. The synthetic datasets were generated using the method of \cite{guyon2003design}. This method  gives us control over the number of informative and redundant features in the generated dataset (Table \ref{tab:dataset}). We expect better performance in our approach when the number of informative features is smaller than the chosen subspace dimension $d$. We also use MNIST \cite{deng2012mnist} and CIFAR10 \cite{cifar} datasets to evaluate the performance of the proposed method on real-world datasets.

\begin{table}[tb]
\small
\begin{center}
\caption{Synthetic Datasets}
\label{tab:dataset}
\begin{tabular}{l|c|c|c|c}
\hline
Dataset & \# training & \# test & \# features & \# Informative \\
\hline\hline
        Dataset 1  & 10000  & 1000 & 64 & 32\\
        Dataset 2  & 10000  & 1000 & 64 & 16\\
        Dataset 3  & 10000  & 1000 & 64 & 8\\
\hline
\end{tabular}
\end{center}
\end{table}

In  experiments on the synthetic datasets, we compare our approach with SQ \cite{wang2016supervised}. We use the same quantizer size as SQ ($\mathcal{C}_k=256$) and fix the subspace dimension $d$ to $16$. Other hyperparameters are chosen in the manner described by \cite{wang2016supervised}. Each comparison is performed for the same code-length and quantizer size. 

In the experiments on real-world datasets, we compare our method with both SQ and PQN. For each experiment, we use the same code length as the baseline. We perform two sets of comparisons. First, we train the model on all of the training set and use all of the test set for evaluations. These experiments show the advantage of using the proposed method under the same evaluation conditions as the baselines. Recently, however, \cite{sablayrolles2016how} showed that it is important to evaluate supervised encoding methods under conditions in which not all classes are known. As such, our second set of tests makes use of the unknown-classes setup suggested by \cite{sablayrolles2016how}, choosing $75\%$ of the classes at random for training on the training set, and testing over the remaining classes. We show that under this condition, we can also outperform existing methods.

\subsection{Experiments}
\subsubsection*{Synthetic Dataset Comparison}
\begin{figure*}[tb]
    \centering
    \subfloat[Dataset 1]{{\includegraphics[width=0.25\linewidth]{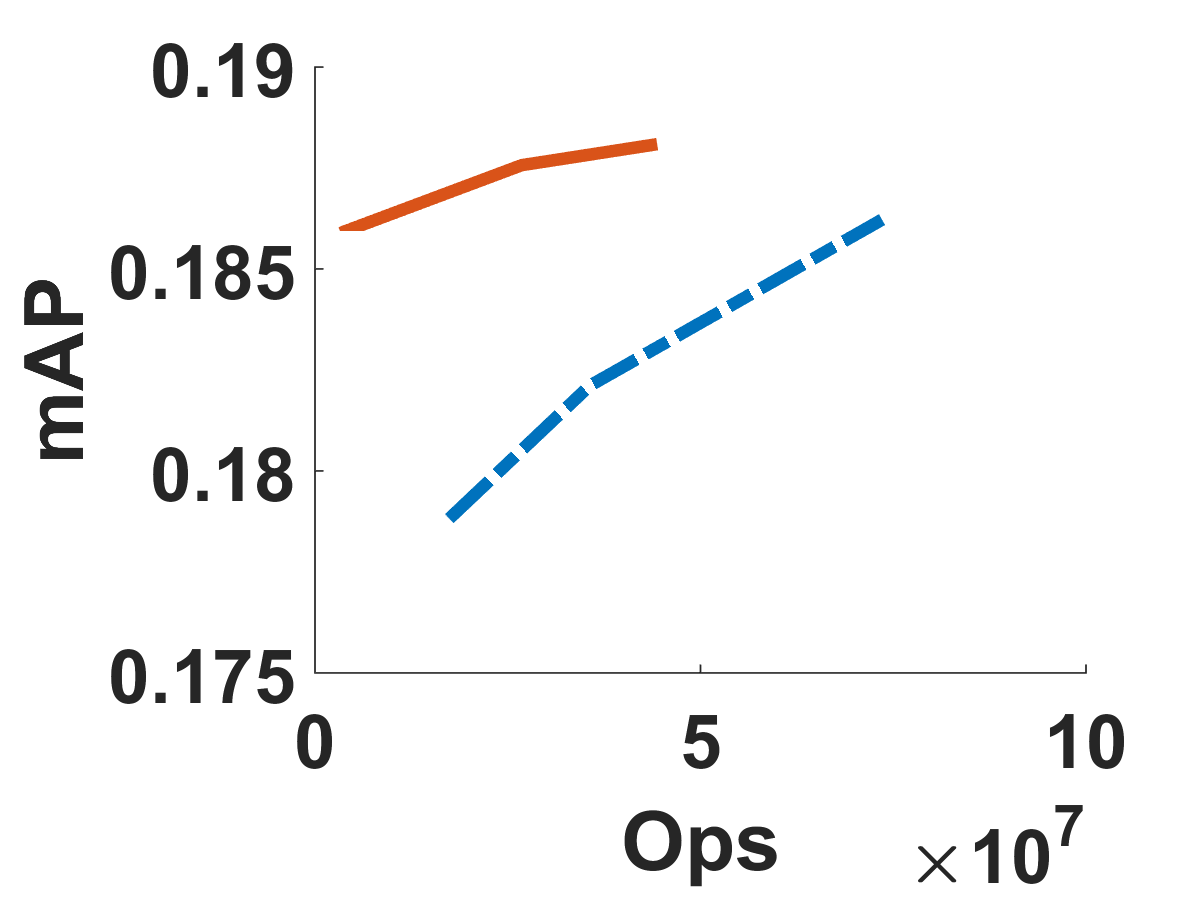} }}%
    \subfloat[Dataset 2]{{\includegraphics[width=0.25\linewidth]{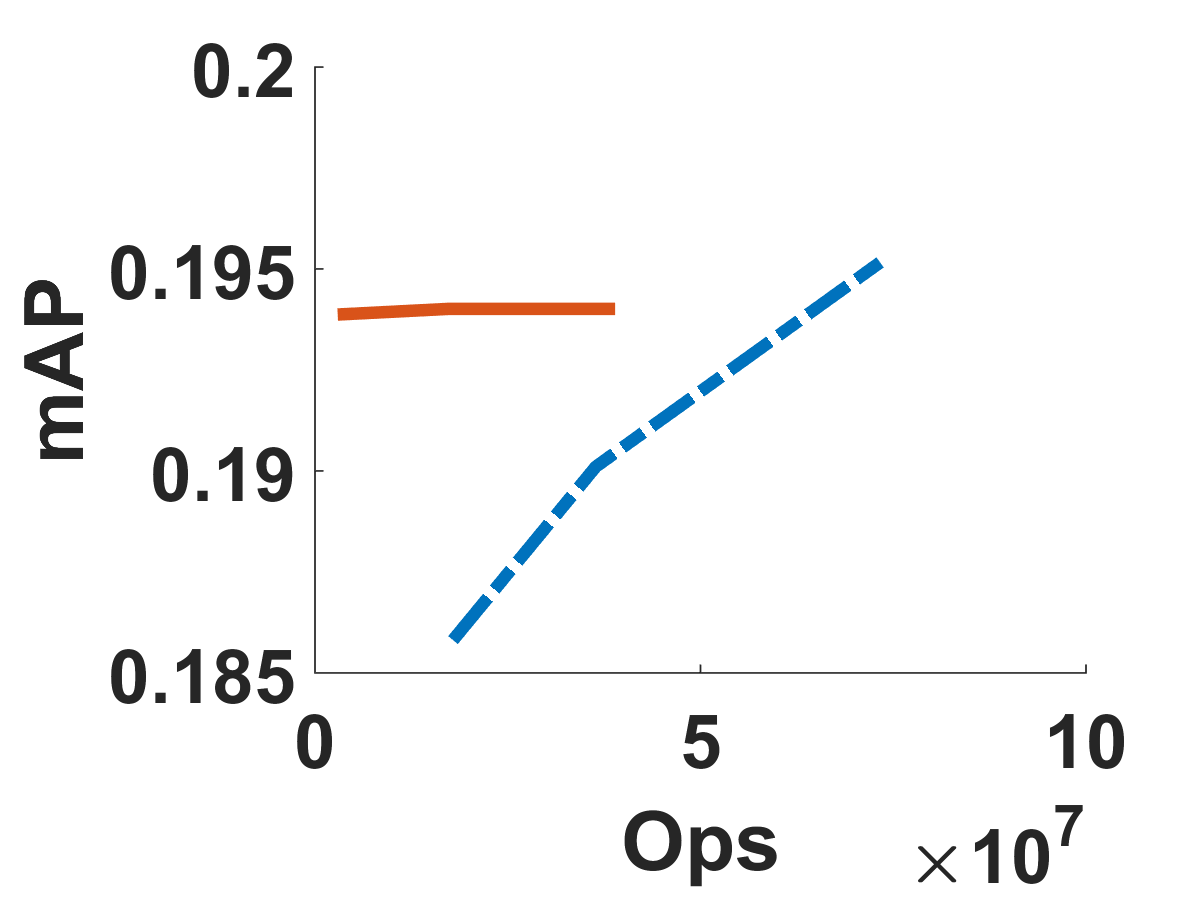} }}%
    \subfloat[Dataset 3]{{\includegraphics[width=0.25\linewidth]{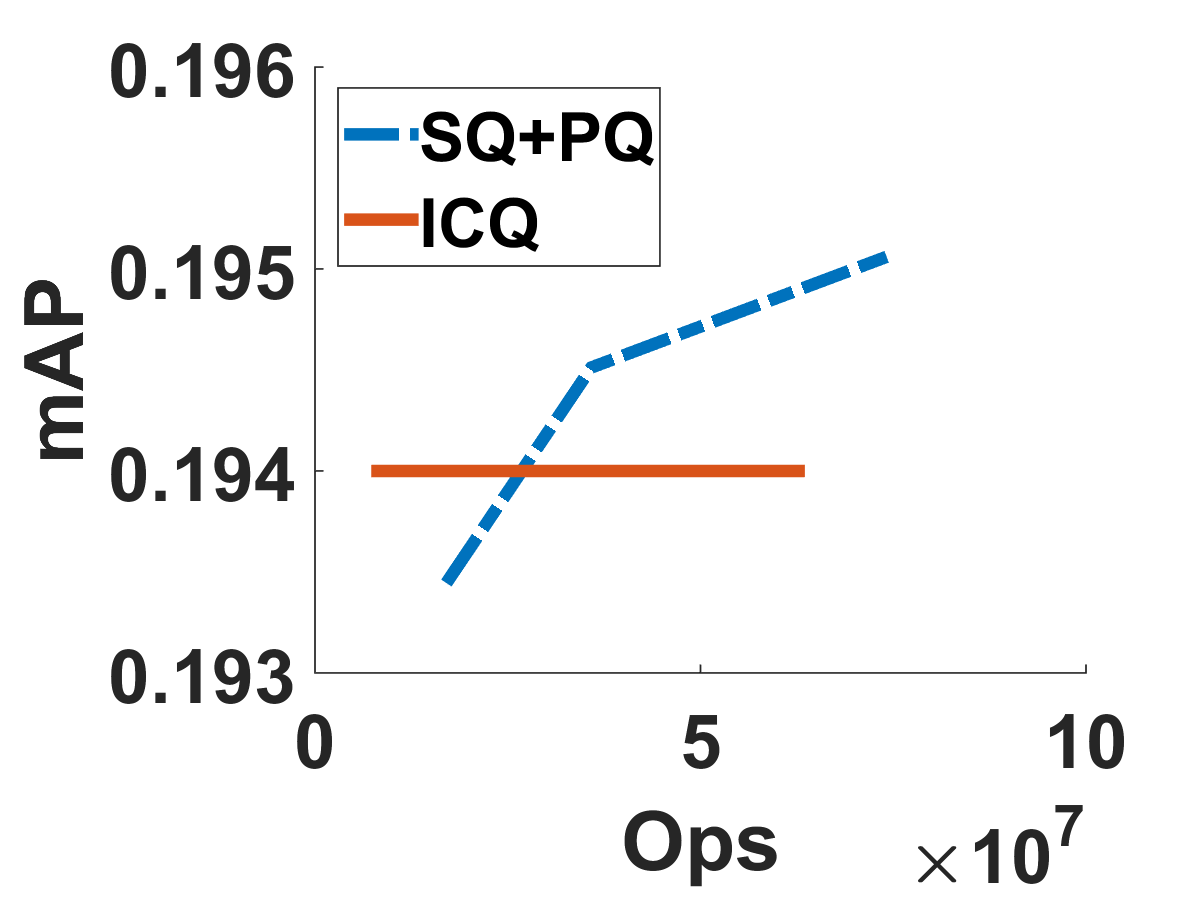} }}%
    \caption{Precision comparison between ICQ and SQ combined with PQ.}%
    \label{fig:sq_pq}
\end{figure*}

\begin{figure*}[tb]
    \centering
    \subfloat[Dataset 1]{{\includegraphics[width=0.25\linewidth]{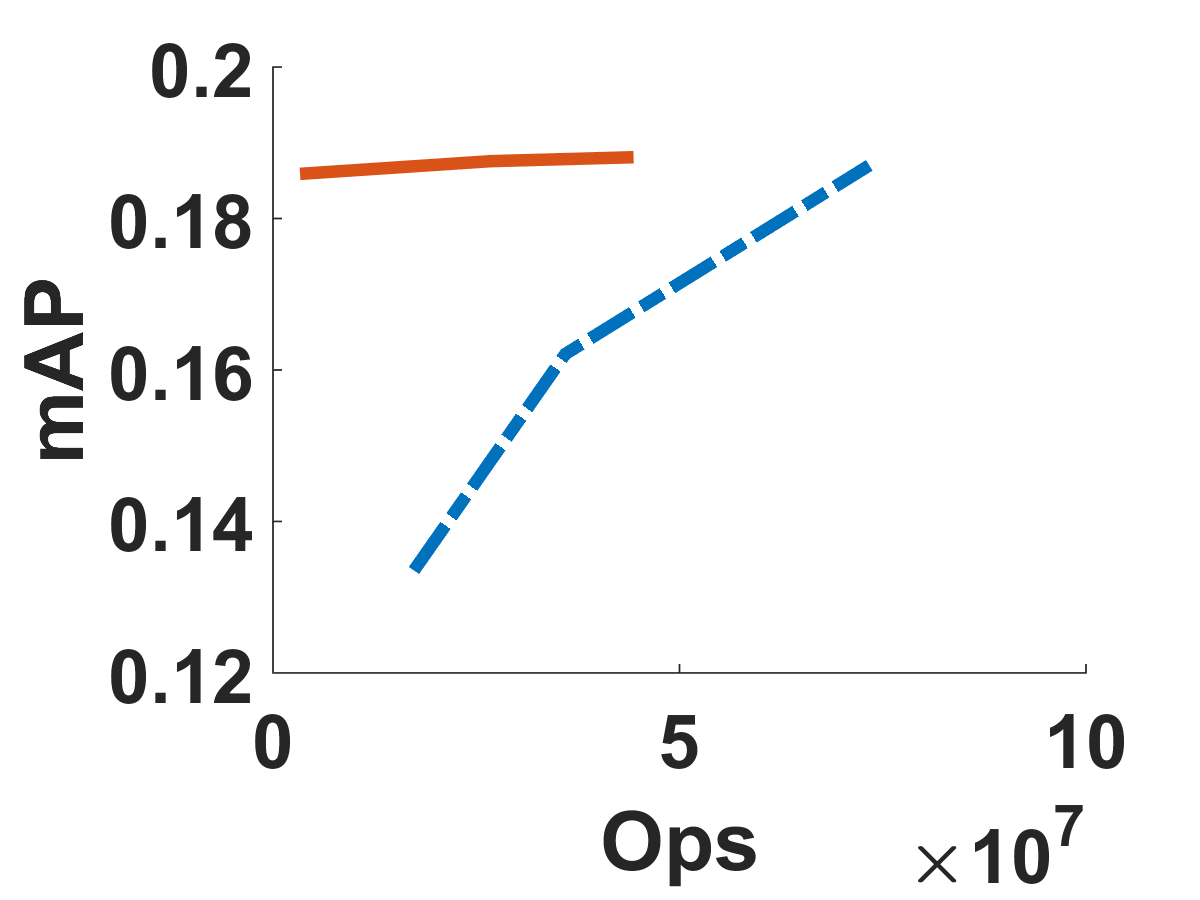} }}%
    \subfloat[Dataset 2]{{\includegraphics[width=0.25\linewidth]{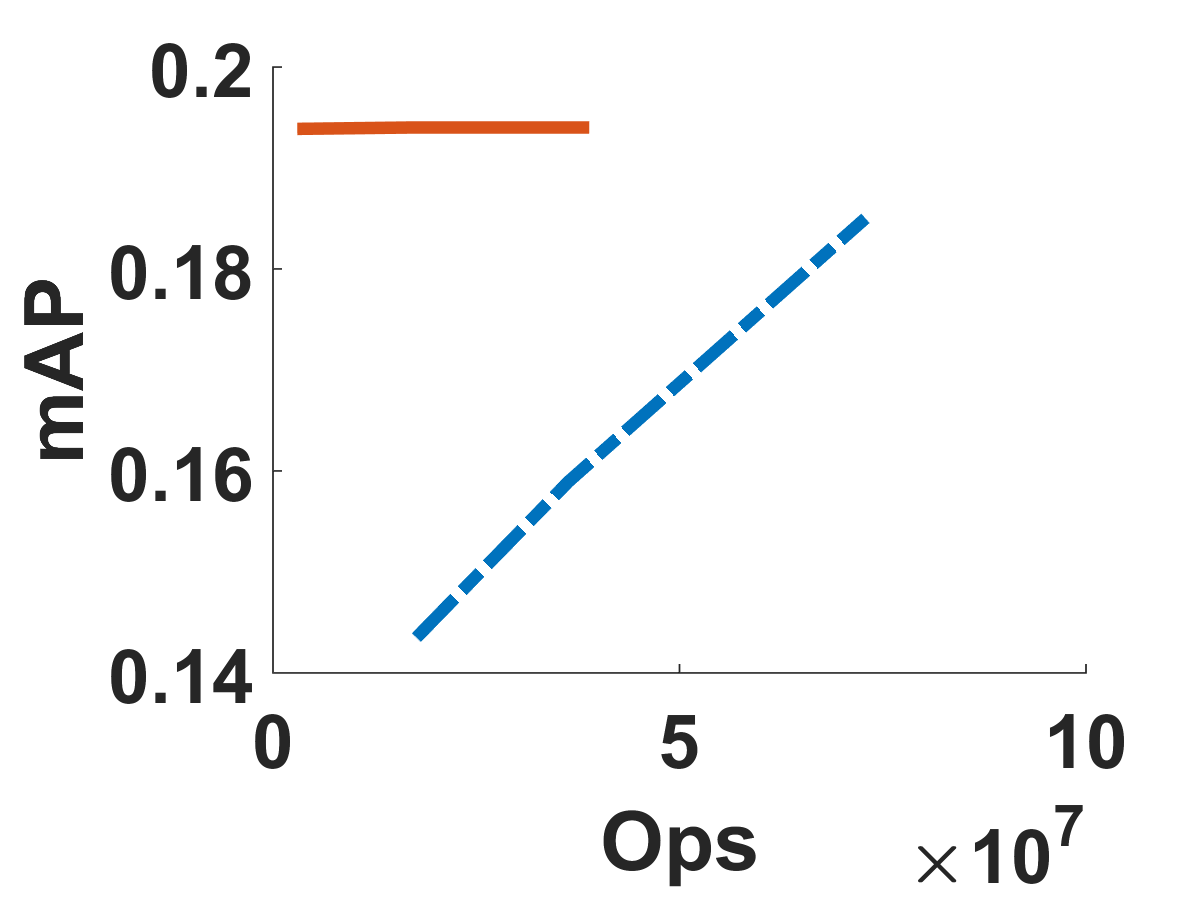} }}%
    \subfloat[Dataset 3]{{\includegraphics[width=0.25\linewidth]{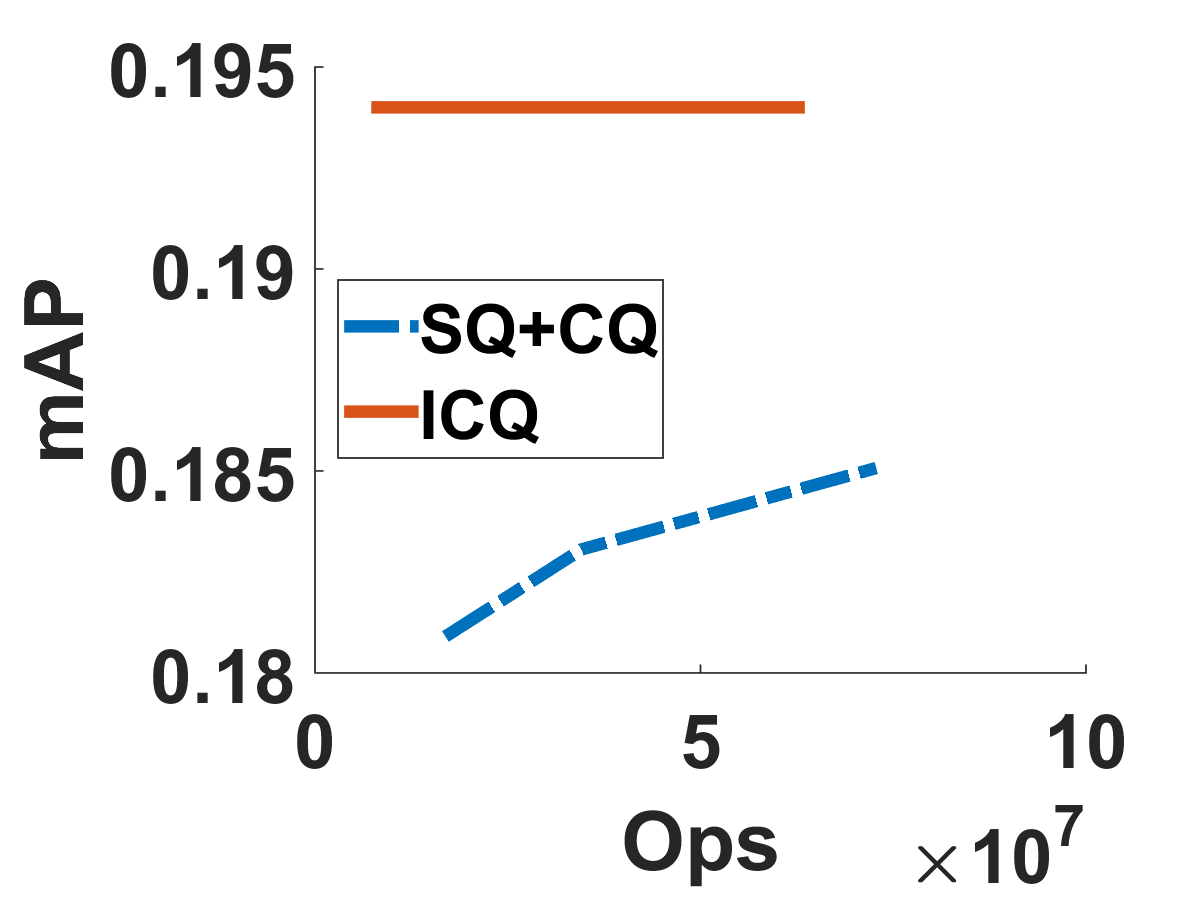} }}%
    \caption{Precision comparison between ICQ and SQ combined with CQ.}%
    \label{fig:sq_cq}
\end{figure*}

For the accuracy test of the synthetic datasets, we compare our approach with SQ both when combined with PQ as well as CQ. Results are shown in Figure~\ref{fig:sq_pq} and Figure~\ref{fig:sq_cq}, respectively. Each point in these diagrams is obtained by first training the coding using one code length, and then averaging the number of operations required to perform a search over the test dataset. 

Figure~\ref{fig:sq_pq} shows that, overall, ICQ is able to perform faster searches and, for the same speed, it achieves higher precisions, due to the efficiency of its approximate distance comparisons. To verify that these improvements are in fact a result of the proposed technique and not the different quantization, we  compared our technique with SQ based on CQ. Results are shown in Figure \ref{fig:sq_cq}. While the SQ$+$CQ produces good results when the number of informative features and the embedding space dimension ($d$) are close (dataset 3), ICQ performs better when there are many informative features due to the high effective dimensionality of the proposed technique.

\subsubsection*{Real-world Dataset Comparison}
For the real datasets, we compare our results with previous works for different numbers of quantizers ($K$). First, similar to the previous experiments, compare ICQ combined with linear mapping with SQ$+$CQ. We also use the same setup to compare ICQ against methods with deep learning embedding methods. Then, we compare ICQ with PQN when both use CNN for embedding. Finally, we show that ICQ outperforms previous method for the case where some classes are excluded during training.

\begin{figure}[tb]%
    \centering
    \subfloat[MNIST: Quantizer Size]{{\label{fig:mnist_size}\includegraphics[width=0.4\linewidth]{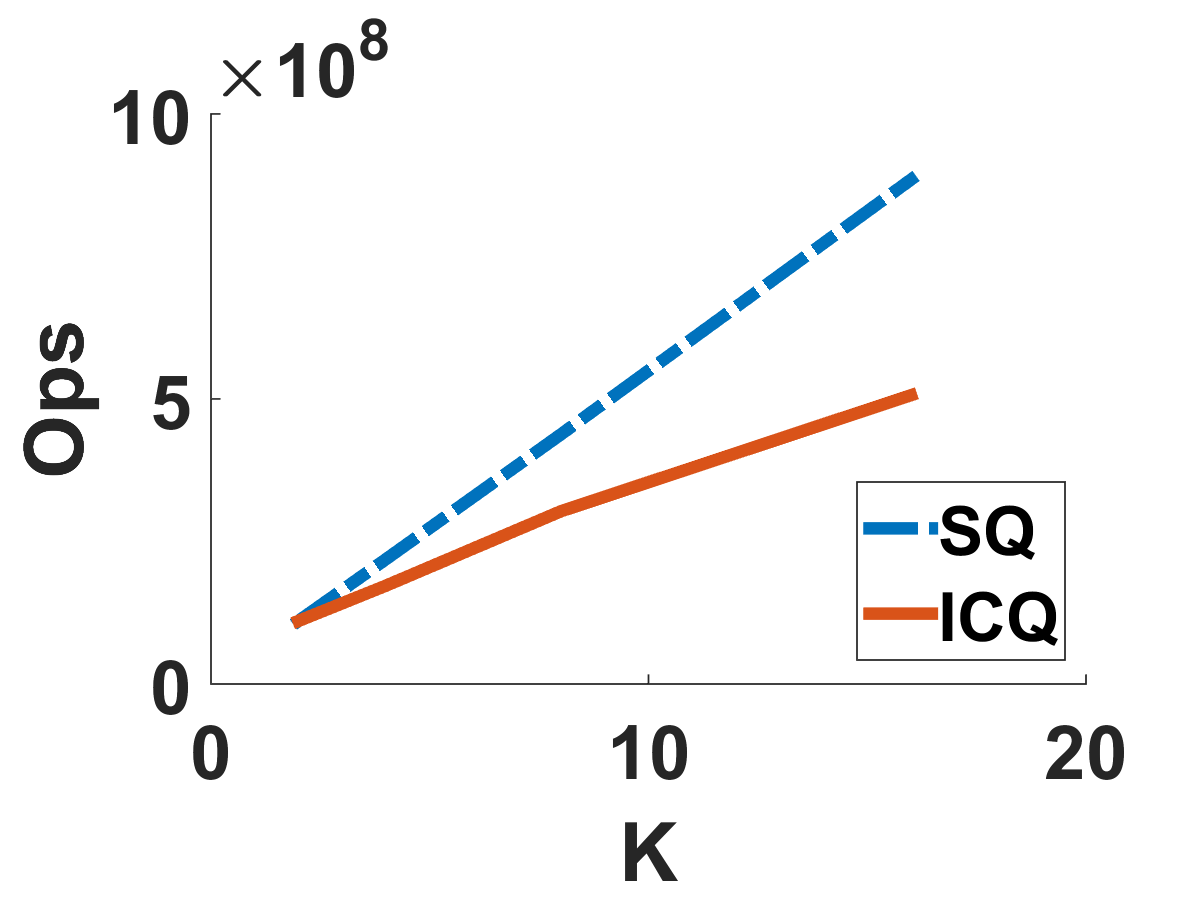} }}%
    \subfloat[MNIST: Precision]{{\label{fig:mnist_precision}\includegraphics[width=0.4\linewidth]{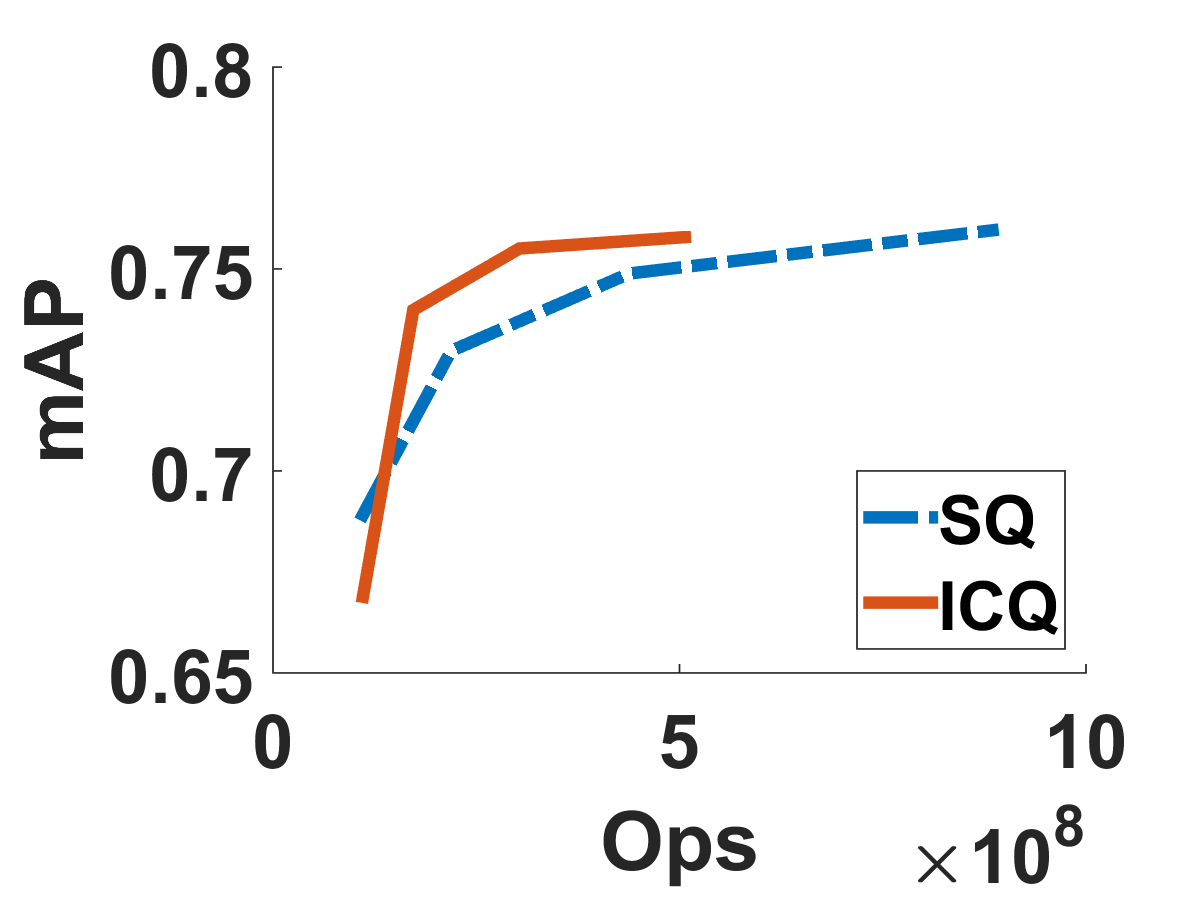} }}\\
    \subfloat[CIFAR:Quantizer Size]{{\label{fig:cifar10_size}\includegraphics[width=0.4\linewidth]{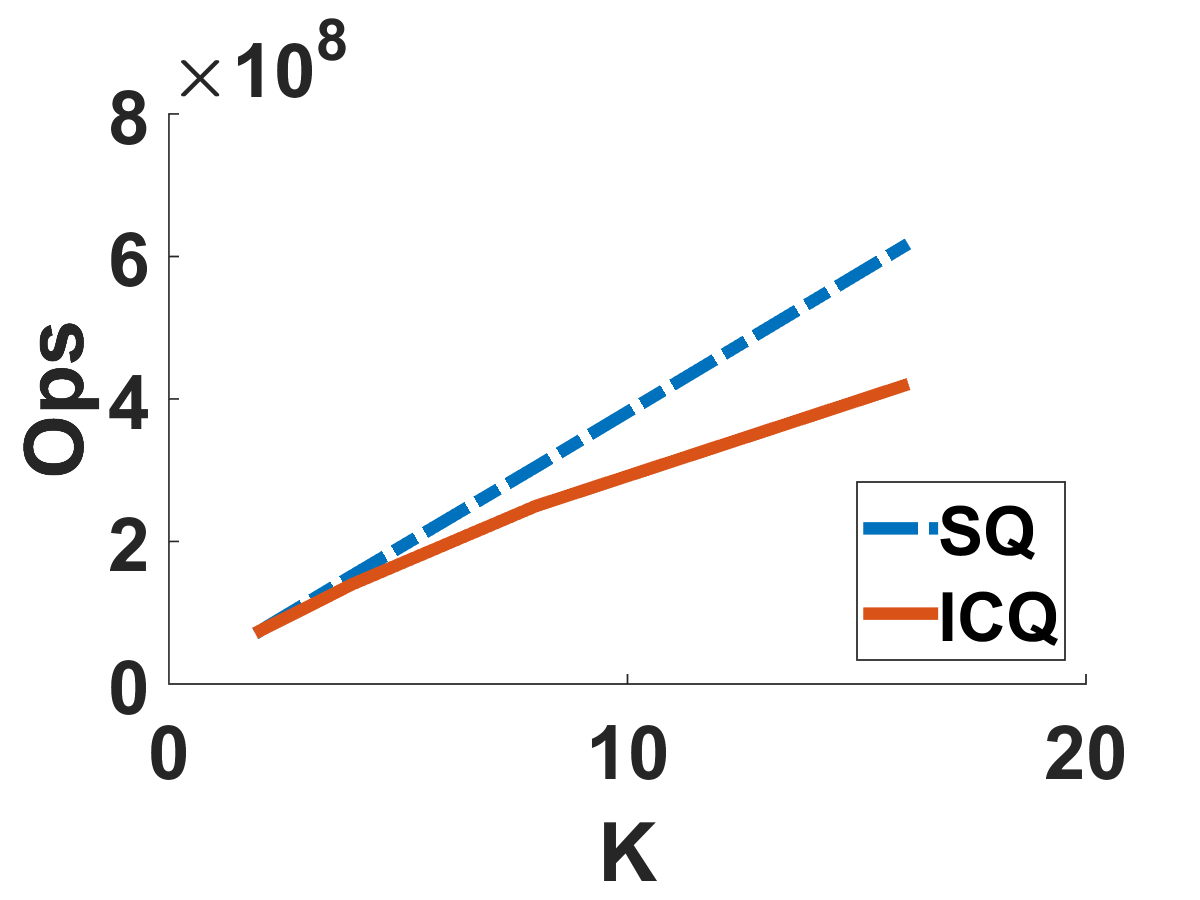} }}
    \subfloat[CIFAR: Precision]{{\label{fig:cifar10_precision}\includegraphics[width=0.4\linewidth]{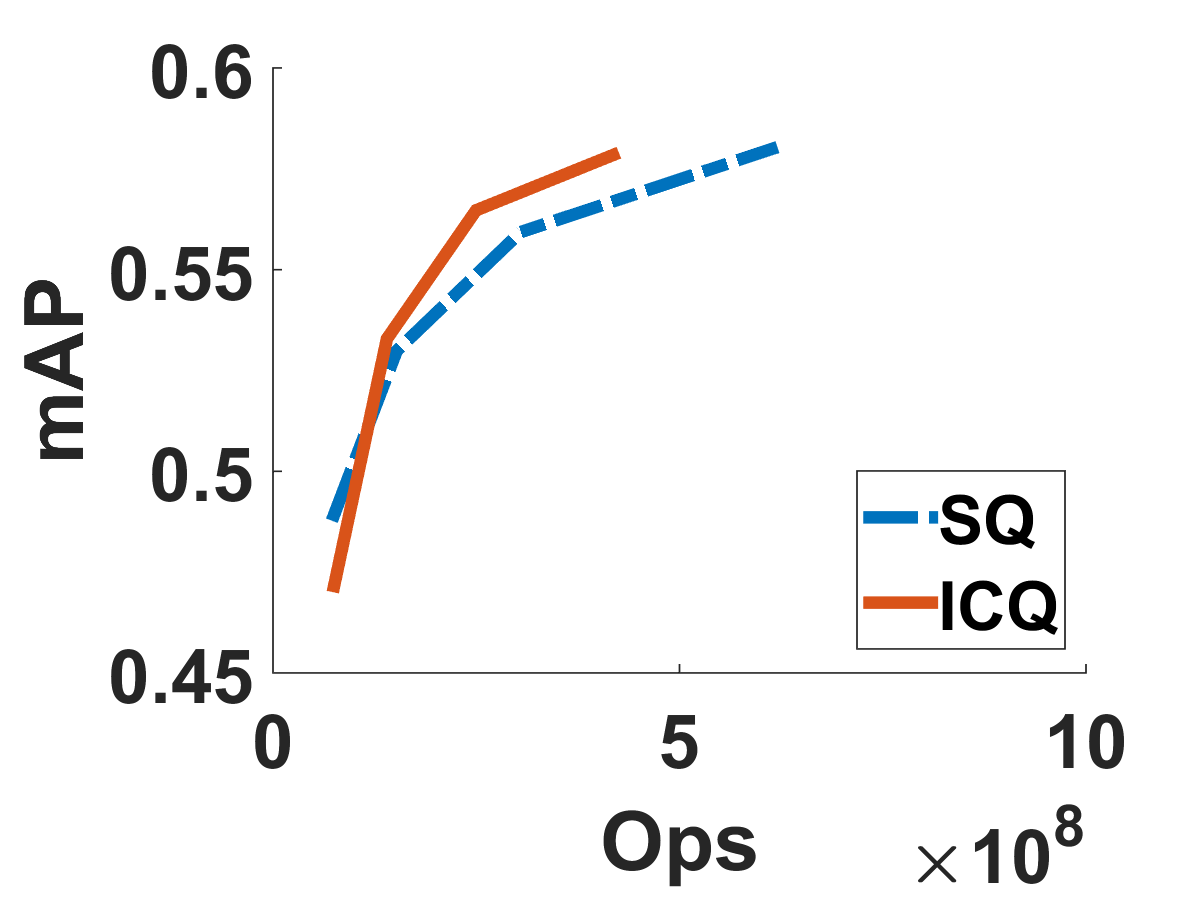} }}%
    \caption{Comparison of ICQ and SQ over MNIST and CIFAR10}%
    \label{fig:icq_sq}
\end{figure}

We have included the results comparing with SQ for the MNIST and CIFAR-10 datasets in Figure \ref{fig:icq_sq}. As Figures \ref{fig:mnist_size} and \ref{fig:cifar10_size} show for $K=2$, both approaches have the same computation load. That is because the proposed approach has to utilize both quantizers to quantize the whole embedding space $\mathbb{R}^{d}$ and thus skips crude distance estimation. Looking at Figures \ref{fig:mnist_precision} and \ref{fig:cifar10_precision}, this case sacrifices significant precision in both ICQ and SQ. Increasing the number of quantizers helps improve precision in both SQ and ICQ, but imposes significantly lower computation costs for ICQ. Figures \ref{fig:mnist_size} and \ref{fig:cifar10_size} show that with more quantizers, the computation cost gap between the two approaches increases and when using $16$ quantizers where precision peaks, ICQ is significantly faster.

\begin{figure}[ht]
    \centering
    \includegraphics[width=0.4\linewidth]{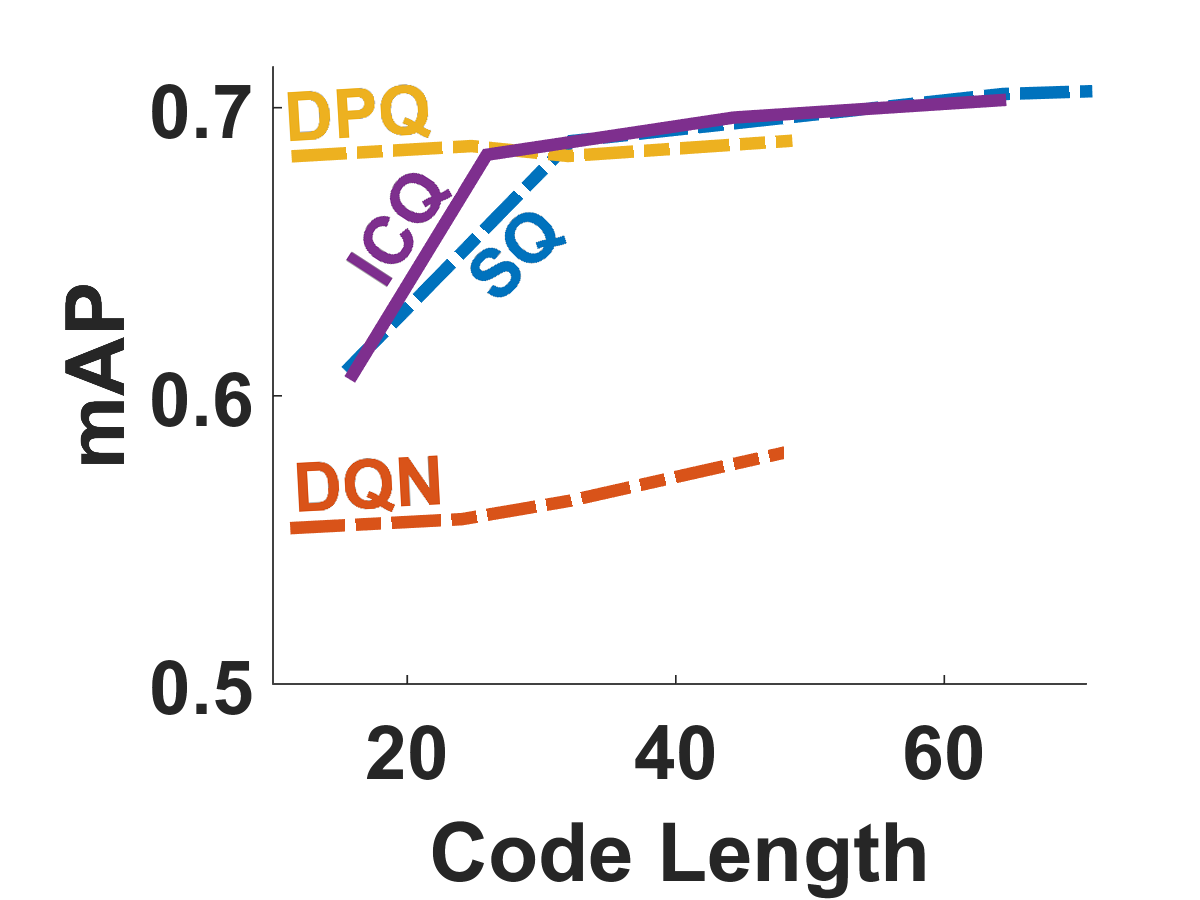}
    \caption{Comparison of ICQ with existing works for different effective code lengths}
    \label{fig:effective}
\end{figure}

Next, we compare ICQ combined with the same linear embedding with several recently proposed quantization methods, and show that even without a sophisticated embedding method, the proposed approach can provide competitive results, improving search speed without shorter codes. Thus, for the sake of a fair comparison between ICQ and existing methods, we compute an effective code length for ICQ as follows: For ICQ using the code length $\ell$, the effective code length $\hat{\ell} \leq \ell$ is chosen to be the code length that SQ would have to use to achieve the same search speed. That is, assuming $\flops_{ICQ@\ell}$ and $\flops_{SQ@\ell}$ are the Average Ops for code length $\ell$ under ICQ and SQ, $\hat{\ell}$ is defined as,
\begin{equation}
    \hat{\ell} = \ell\times\frac{\flops_{ICQ@\ell}}{\flops_{SQ@\ell}}
\end{equation}
We compare the MAP over the CIFAR-10 dataset achieved by ICQ for different effective code lengths against Deep Quantization Network (DQN) \cite{cao2016deep} and Deep Product Quantization (DPQ) \cite{klein2017defense} in Figure \ref{fig:effective}. ICQ can outperforms both SQ and DQN in effective code length. Further, it outperforms DPQ in large code lengths with a simpler embedding method.

\begin{figure}[tb]
    \centering
    \subfloat[MNIST]{{\label{fig:cifar_size}\includegraphics[width=0.4\linewidth]{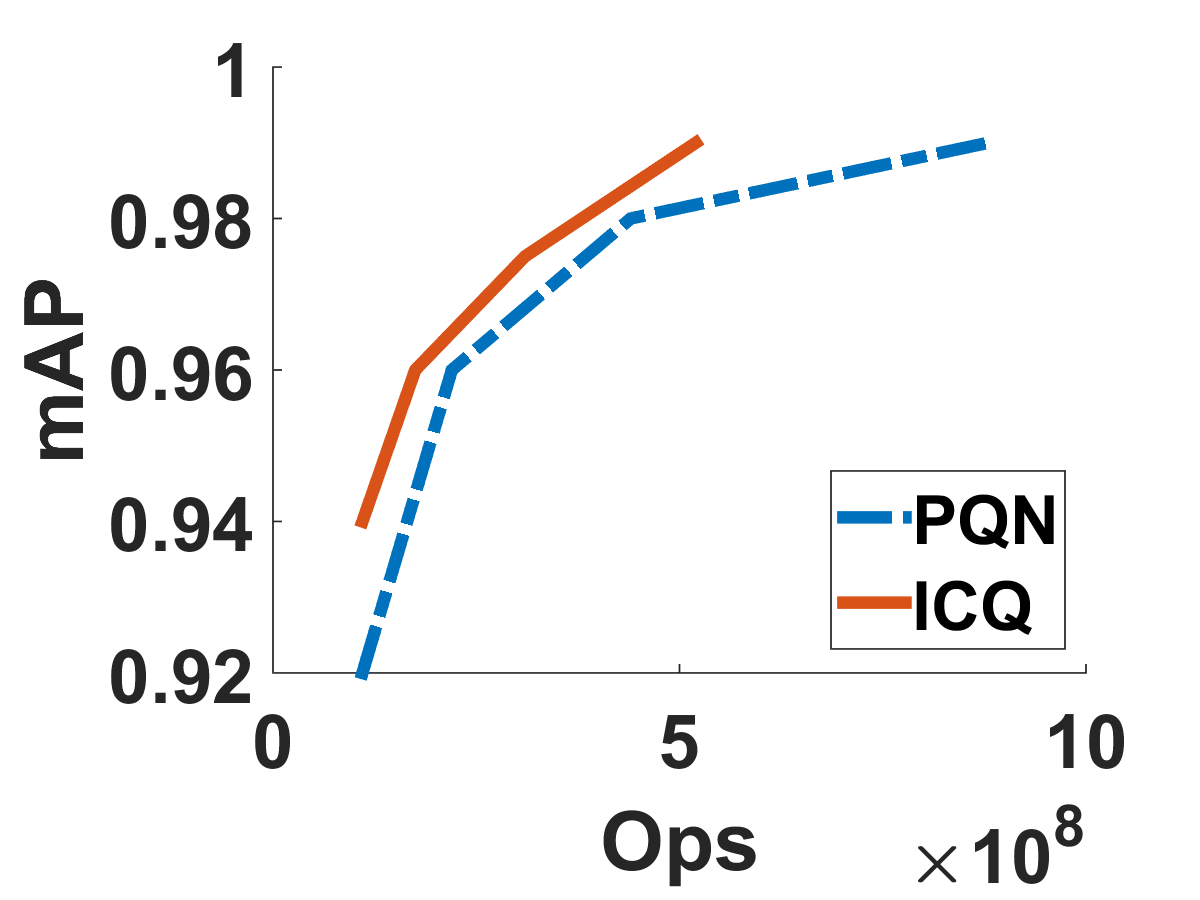} }}
    \subfloat[CIFAR-10]{{\label{fig:cifar_precision}\includegraphics[width=0.4\linewidth]{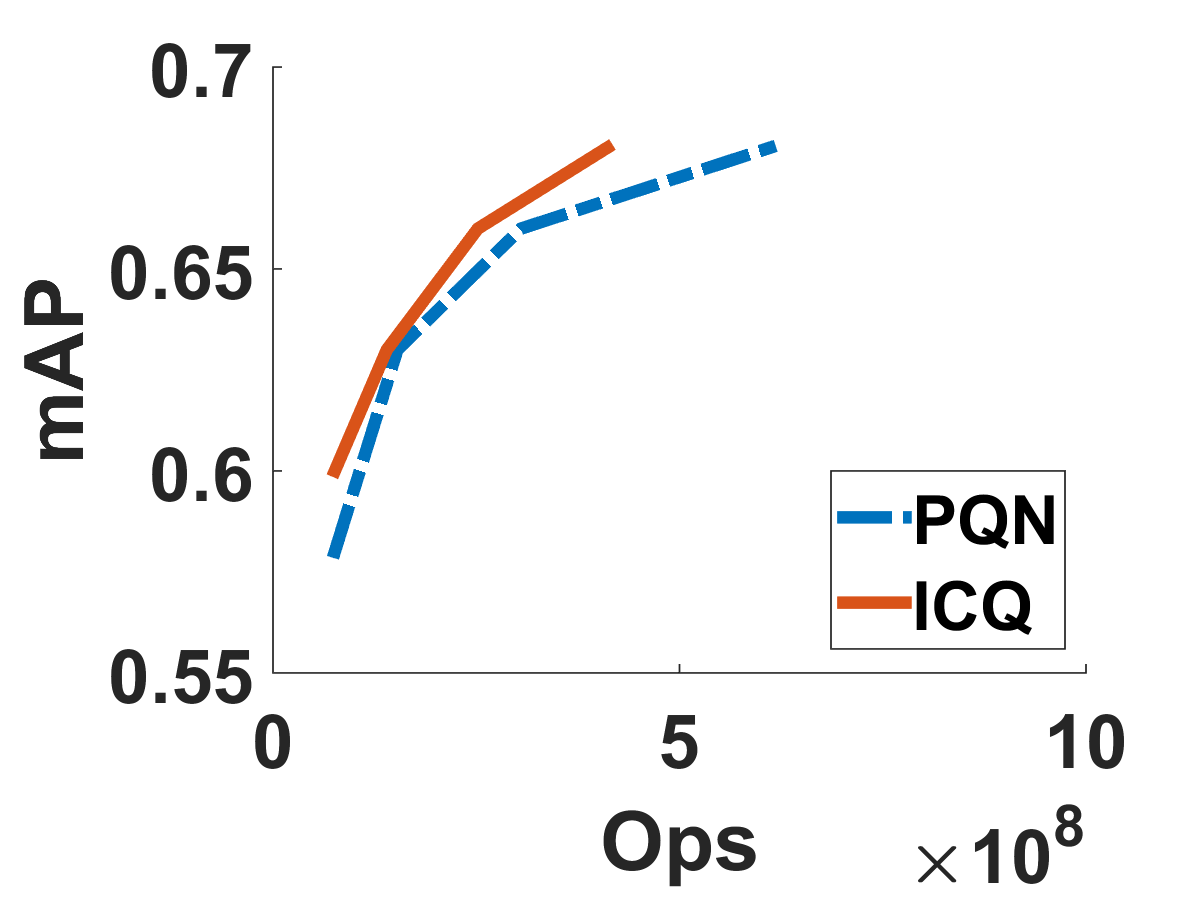} }}%
    \caption{Comparison of ICQ and PQN}%
    \label{fig:pqn}
\end{figure}

Finally, we compare our method with PQN \cite{yu2018product}. For this comparison, we use a CNN for embedding, similar to PQN, and replace the quantization method, PQ, with ICQ. We use LeNet \cite{lecun1998gradient} for MNIST and AlexNet \cite{krizhevsky2012imagenet} for CIFAR-10 with 512 and 1024-dimensional embeddings, respectively. Since, PQN trains on triplets of inputs, we randomly generate 400K triplets in each case. We perform all comparisons for the same code length. Results are shown in Figure \ref{fig:pqn} for MNIST and CIFAR-10, demonstrating the advantage of the proposed method over the state-of-the-art. In contrast to the comparison with SQ, we have an advantage over PQN even for smaller code lengths. That is because ICQ, similar to CQ which was used by SQ, allows quantizers to be dense in the embedding space. On the other hand, PQ quantizers are sparse, so they normally incur higher quantization errors. More importantly, $\mathcal{K}$ is increased, we can see for the same code lengths, the proposed method performs searches faster, which is the result of the two-step search operations. This advantage persists even if we combine PQN with CQ. In summary, the proposed method consistently outperforms PQN.

\subsubsection*{Comparison over unseen classes}
\begin{figure}[tb]
\centering
\subfloat[MNIST]{%
\label{fig:first}%
\includegraphics[width=0.4\linewidth]{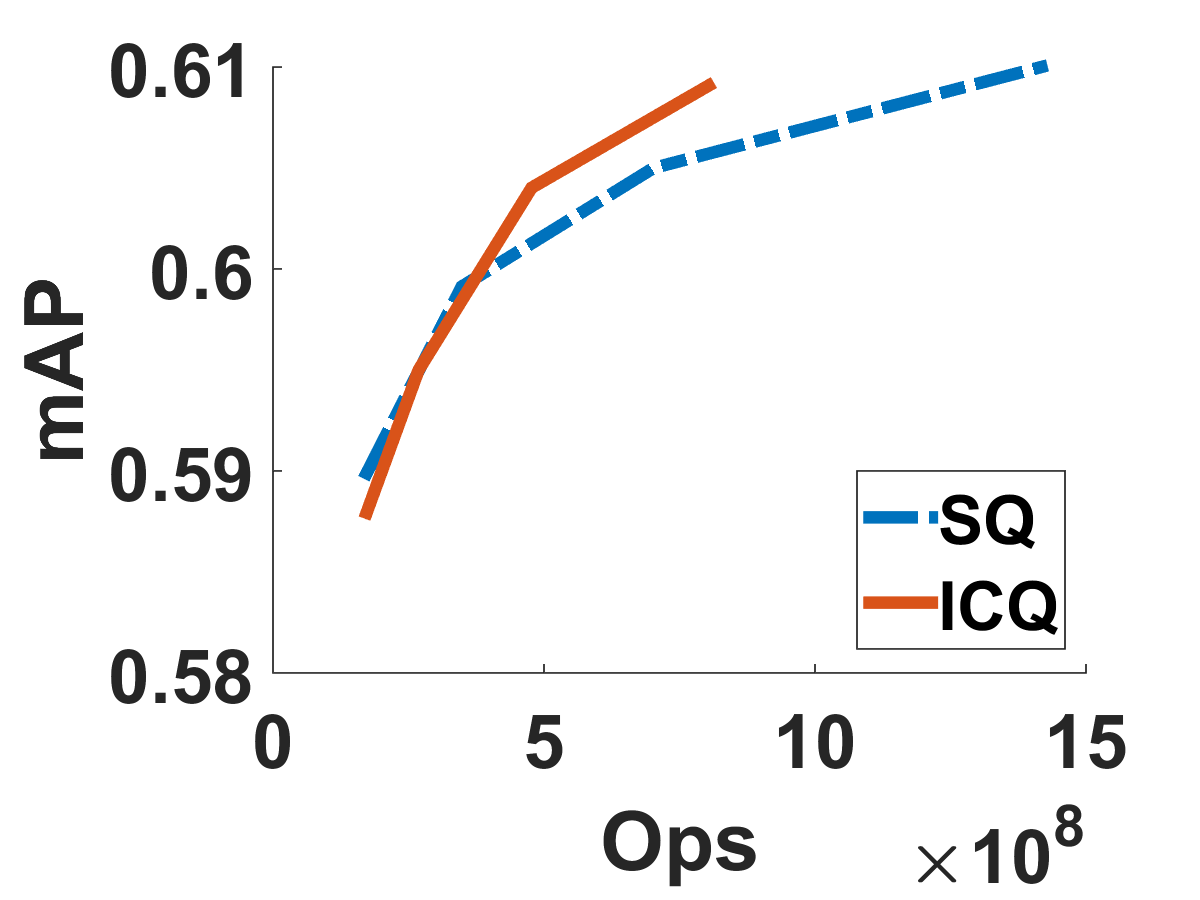}}%
~\enskip
\subfloat[CIFAR-10]{%
\label{fig:second}%
\includegraphics[width=0.4\linewidth]{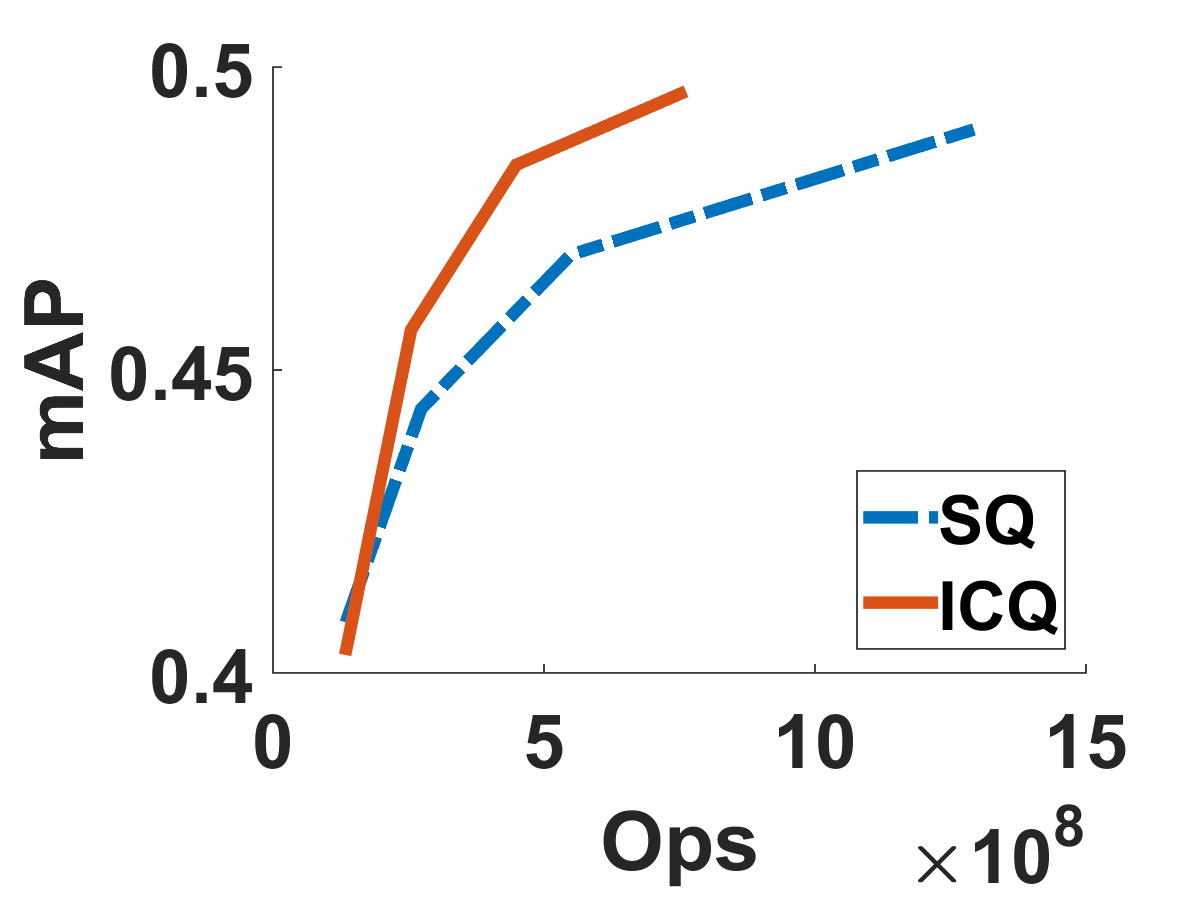}}%
\caption{Comparison of ICQ and SQ over unseen classes}
\label{fig:unseen}
\end{figure}

We use the methodology proposed by \cite{sablayrolles2016how} to further evaluate ICQ combined with linear mapping, and compare the results with SQ. In these experiments we leave out three randomly selected classes during training, and report search accuracy over these three classes. The results of these experiments for coding with different encoding lengths over MNIST and CIFAR10 are shown in Figure \ref{fig:unseen}. We see that ICQ also outperforms the baseline in classifying unseen classes.
\section{Conclusion}
We proposed Interleaved Composite Quantization (ICQ), a method which enables fast similarity search. The proposed technique learns a prior on the distribution of the variance of the dataset, and by incorporating this prior into learning the quantizers, it is able to perform fast approximate distance calculations. This approximation is sufficient for similarity search in most cases; the computationally heavy exact comparisons can be reserved to disambiguate corner cases. We tested the ICQ over several datasets and showed that it can consistently outperform the current state of the art.
%\section{Acknowledgement}
%\input{src/s6_acknowledgement}

{\small
\bibliographystyle{ieee_fullname}
\bibliography{ms}
}

\end{document}